\documentclass{article}

\usepackage{microtype}
\usepackage{graphicx}
\usepackage{subcaption}
\usepackage{booktabs} % for professional tables

\usepackage{hyperref}

\usepackage[preprint]{icml2026}

\usepackage{amsmath}
\usepackage{amssymb}
\usepackage{mathtools}
\usepackage{amsthm}

\usepackage[capitalize,noabbrev]{cleveref}

\usepackage{enumitem}
\usepackage{stfloats}
 \usepackage{tikz}

\theoremstyle{plain}
\newtheorem{theorem}{Theorem}[section]

\theoremstyle{definition}
\newtheorem{definition}[theorem]{Definition}

\theoremstyle{remark}

\usepackage[textsize=tiny]{todonotes}

\icmltitlerunning{Selecting Optimal Variable Order in Autoregressive Ising Models}

\DeclareMathOperator*{\argmin}{arg\,min}

\begin{document}

\twocolumn[
  \icmltitle{Selecting Optimal Variable Order in Autoregressive Ising Models}



  \icmlsetsymbol{equal}{*}

  \begin{icmlauthorlist}
    \icmlauthor{Shiba Biswal}{yyy}
    \icmlauthor{Marc Vuffray}{yyy}
    \icmlauthor{Andrey Y. Lokhov}{yyy}
  \end{icmlauthorlist}

  \icmlaffiliation{yyy}{Theoretical Division, Los Alamos National Laboratory, Los Alamos, NM 87545}

  \icmlcorrespondingauthor{Shiba Biswal}{sbiswal@lanl.gov}

  \icmlkeywords{Machine Learning, ICML}

  \vskip 0.3in
]



\printAffiliationsAndNotice{}  % no special notice (required even if empty)

\begin{abstract}
    Autoregressive models enable tractable sampling from learned probability distributions, but their performance critically depends on the variable ordering used in the factorization via complexities of the resulting conditional distributions. We propose to learn the Markov random field describing the underlying data, and use the inferred graphical model structure to construct optimized variable orderings. We illustrate our approach on two-dimensional image-like models where a structure-aware ordering leads to restricted conditioning sets, thereby reducing model complexity. Numerical experiments on Ising models with discrete data demonstrate that graph-informed orderings yield higher-fidelity generated samples compared to naive variable orderings.
\end{abstract}

\section{Introduction}
Autoregressive models are powerful decompositions widely used in modern AI architectures for producing exact samples from a learned representation of the data. Similar to the procedure used in a Bayesian network, new samples are generated using \textit{ancestral sampling}: variables are visited in a topological order, and each variable $x_i$ is conditioned on the previously sampled ``parent'' variables ${\bf x}_{<i}$, yielding the factorization
\begin{equation}
\label{eq:IntroCP}
 p(x) = \prod_i p(x_i \mid {\bf x}_{<i}).   
\end{equation}

Each conditional distribution is learned from training data in advance. In practice, the ordering in which variables are traversed is rarely optimized and is typically chosen arbitrarily or dictated by a natural sequence in the data (e.g., word order in text or a fixed pixel traversal in images). However, not all orderings are equally effective: different orderings can induce conditional distributions of vastly different complexity. This in turn affects sample accuracy: good orderings simplify conditionals and reduce error propagation, while poor orderings force the model to learn unnecessarily complex dependencies. 

In the case where the joint probability distribution is represented as a Markov Random Field (MRF), this structure can be leveraged to construct optimized variable orderings. In general, the conditional $p(x_i \mid {\bf x}_{<i})$ can be a complex distribution if it depends on all previously sampled variables. However, using the Markov property, the effective conditioning set is reduced to a subset of neighboring nodes in the underlying undirected graphical model (see \Cref{fig:example} below for a concrete example of this mechanism). If the MRF structure is not a priori known, it can be first learned from data.

In this work, we construct optimized variable orderings that leverages the structure of the MRF associated with the data-generating distribution. For concreteness, we focus on a popular class of two-dimensional lattice models that are widespread in physical systems and in probabilistic image models. Our approach selects variable orderings that minimize the size of the conditioning set for each conditional distribution and exploit the decay of correlations with lattice distance that is characteristic of many probability densities. Through numerical experiments on discrete distributions on binary variables,
% where conditionals are learned using the Interaction Screening family of estimators with explicit energy-based parametrization \cite{vuffray2016interaction},
we demonstrate that MRF-informed traversals produce higher-fidelity samples than naive or arbitrary orderings, for a fixed and finite number of training samples.

\paragraph{Related work.} Several works employ neural networks to learn the conditional distributions in the autoregressive factorization~\eqref{eq:IntroCP} and explicitly note that the choice of variable ordering affects the complexity of the resulting conditionals. For example, the Neural Autoregressive Distribution Estimator (NADE)~\cite{uria2016neural} factors the joint distribution into a sequence of conditional distributions and models each conditional using a neural network. Building on this idea, the Masked Autoencoder for Distribution Estimation (MADE)~\cite{germain2015made} enforces the autoregressive structure through masking in feedforward networks, enabling efficient learning of all conditionals within a single model while respecting a prescribed variable ordering. Both NADE and MADE explicitly acknowledge that the choice of ordering strongly influences the complexity of the learned conditional distributions and the quality of generated samples. In the context of Ising models, MADE architecture has been applied to fully-connected models in \cite{del2025performance}.

Related issues arise in autoregressive generation of structured objects, particularly graphs, where no canonical ordering exists. Autoregressive graph generation methods such as GraphRNN \cite{you2018graphrnn} and GraphAF \cite{shi2020graphaf} rely on heuristic traversal strategies (e.g., breadth-first or depth-first search), and the chosen ordering has been shown to significantly affect model expressivity, stability, and sample quality. More recently, Biazzo et al.~\cite{biazzo2024sparse} proposed an autoregressive neural network architecture specifically tailored to sparse two-body interacting spin systems, incorporating structural information from the underlying Hamiltonian. In a related direction, Bia{\l}as et al.~\cite{bialas2022hierarchical} introduced hierarchical autoregressive neural networks for modeling high-dimensional probability distributions arising in lattice field theories and statistical mechanics, such as the Ising model, with improved scalability for large systems. Masked Autoregressive Flows (MAFs)~\cite{papamakarios2017masked} further combine autoregressive conditionals with invertible transformations to enable exact likelihood evaluation and sampling, often mitigating sensitivity to a fixed ordering by interleaving multiple autoregressive layers with permutations of the variables.

Concerns about variable ordering also arise prominently in Bayesian network structure learning, where the ordering of variables can significantly affect the learned graph, model complexity, and predictive performance. For example, \citet{kitson2024impact} analyze how different orderings impact score-based structure learning, while subsequent work~\cite{kitson2024eliminating} addresses the instability induced by variable order choices in greedy learning algorithms. Although these studies focus on structure learning rather than autoregressive sampling, they reinforce the broader observation that variable ordering is a fundamental modeling choice in probabilistic systems.

Any-Order Autoregressive Models (AO-ARMs)~\cite{germain2015anyorder} have been proposed to mitigate sensitivity to a fixed ordering by training a single model that supports multiple or arbitrary permutations. Such approaches, including orderless variants of NADE, aim to make the learned distribution robust to ordering by effectively averaging over many factorizations. In contrast to these order-agnostic approaches, our work explicitly leverages the Markov random field structure underlying scientific data to construct orderings that minimize conditional complexity and improve the fidelity of autoregressive sampling.

In this work, we specifically focus on autoregressive models in the context of binary random variables. We evaluate the efficacy of different variable orderings by learning conditional distributions from data and resampling to evaluate the performance of the learned model. The techniques for learning such discrete conditional distributions have been developed and analyzed in the graphical model learning literature \cite{klivans2017learning, vuffray2020efficient}.   

\section{Problem and Methods}

\subsection{Problem statement: autoregressive decompositions and variable ordering}
Any joint distribution $p$ over $N$ variables admits an \textit{autoregressive decomposition} of the form
\begin{align}
\begin{aligned}
        p(x) = p(x_1) \; p(x_2|x_1) \; p(x_3|x_1,x_2) \ldots \\ p(x_N|p(x_1,\ldots, x_{N-1}),
\end{aligned}
\end{align}
where $x = (x_1, \ldots, x_N)$, which is a simple consequence of a chain rule. In practice, the conditional probabilities in the decomposition are not known, but can be learned from data. Such a representation of the joint density $p(x)$ is attractive because exact sampling from $p(x)$ reduces to sampling each one-dimensional conditional distribution. Since each step involves sampling a single variable, the total computational cost of generating one sample is $\mathcal{O}(N)$. This is the reason why probabilistic autoregressive models are widespread in applications, including large language models such as GPT-3 and GPT-NeoX-20B \cite{brown2020language, black2022gpt}.

Any variable order can be used to construct an autoregressive decomposition. More generally, for any permutation of nodes $\sigma(1), \ldots, \sigma(N)$ of the node set $\mathcal{V}$ the distribution $p$ can be decomposed as
\begin{align}
\label{eq:general_decomposition}
\begin{aligned}
p(x) = p(x_{\sigma(1)}) \; p(x_{\sigma(2)} | x_{\sigma(1)}) \; p(x_{\sigma(3)} | x_{\sigma(1)}, x_{\sigma(2)}) \\ 
\ldots p(x_{\sigma(N)} | x_{\sigma(1)},\ldots, x_{\sigma(N-1)}).
\end{aligned}
\end{align}
Different permutations of nodes can have a significant impact on sampling quality as they affect the complexity of the conditionals in the decomposition. Indeed, learning the conditional distributions accurately, particularly when the conditioning sets are large, may pose a significant challenge. In this work, we address the following problem: \emph{can we use the structure of the distribution $p(x)$ to select variable ordering leading to conditionals of lower complexity}? We put forward and validate a hypothesis that a selection based on this criterion should generally lead to better-quality autoregressive models. 

For concreteness, we will focus on the specific case of binary models with pairwise interactions, known as Ising models. Ising model is defined on an undirected graph $\mathcal{G}=(\mathcal{V},\mathcal{E})$, where $\mathcal{V} = \{1,\ldots,N\}$ denotes the set of nodes and $\mathcal{E} \subseteq \mathcal{V}\times \mathcal{V}$ denotes the set of undirected edges. Let $x_i \in \{-1,1\}$ denote the spin associated with node $i$, and let the vector $x = (x_1, \ldots, x_N)$ denote the configurations of all spins. The positive probability distribution of the Ising model is given by
\begin{equation}
\label{eq:IsingMod}
    p(x) = \frac{1}{Z} \exp \left( \sum_{i\in \mathcal{V}} \theta_i \,x_i + \sum_{(i,j)\in \mathcal{E}} \theta_{i,j} \,x_i x_j  \right)
\end{equation}
where $\theta_i$ and $\theta_{i,j}$ are (finite) parameters corresponding to the external magnetic fields and pairwise interactions, respectively, and $Z$ is the partition function (normalization constant).

% Since the Ising model is a log-linear Markov random field, its joint distribution belongs to the exponential family; conditioning on a subset of variables preserves this structure, yielding conditional distributions that are themselves exponential-family models, possibly with induced higher-order interactions.
Ising model is a Markov random field, meaning that conditioned on a set of neighbors, any node in the graph is independent of the rest of the nodes. Conditional distributions appearing in an autoregressive decomposition admit an exponential-family form, but possibly with induced higher-order interactions. In what follows, we use the fact that any positive distribution of $n$ random binary variables can be represented as an exponential-family distribution with a polynomial energy function of order at most $n$. For instance, under a given ordering $\sigma$, the first few conditional distributions are
\begin{align}
\label{eq:CondProb}
\begin{aligned}
&p(x_{\sigma(1)}) = \frac{1}{Z_{\sigma(1)}} \exp\!\left( \tilde{\theta}_{\sigma(1)}\, x_{\sigma(1)} \right), \\
&p(x_{\sigma(2)} \mid x_{\sigma(1)}) = \frac{1}{Z_{\sigma(2)}}
\exp\!\left(\tilde{\theta}_{\sigma(2)}\, x_{\sigma(2)} \right. \\
&\hspace{0.5cm} + \left. \tilde{\theta}_{\sigma(1),\sigma(2)}\, x_{\sigma(1)} x_{\sigma(2)}
\right), \\
&p(x_{\sigma(3)} \mid x_{\sigma(1)}, x_{\sigma(2)}) = \frac{1}{Z_{\sigma(3)}} 
\exp\! \left( \tilde{\theta}_{\sigma(3)}\, x_{\sigma(3)}  \right. \\
&\hspace{0.5cm}+ \tilde{\theta}_{\sigma(1),\sigma(3)}\, x_{\sigma(1)} x_{\sigma(3)} 
+ \tilde{\theta}_{\sigma(2),\sigma(3)}\, x_{\sigma(2)} x_{\sigma(3)} \\
&\hspace{0.5cm}+ \left. \tilde{\theta}_{\sigma(1),\sigma(2),\sigma(3)}\, x_{\sigma(1)} x_{\sigma(2)} x_{\sigma(3)} \right),
\end{aligned}
\end{align}
and so on. Here $Z_{\sigma(\cdot)}$ denotes the appropriate conditional normalization constant, and the coefficients $\tilde{\theta}$ are effective parameters induced by the autoregressive ordering. In general, higher-order interaction terms appear in the conditional distributions even when the original Ising model contains only pairwise interactions.

\subsection{Conditional independence and parent sets}

As the size of the conditioning set grows, the number of effective parameters increases rapidly, leading to a corresponding increase in learning complexity. However, the underlying graph structure $\mathcal{E}$ can be exploited to restrict the conditioning set, referred to here as the parent nodes $\operatorname{Par}(\cdot): \mathcal{V} \to E \subseteq \mathcal{V}$, thereby controlling the complexity of the learned conditionals. Towards this goal, we first define a path from $u \in \mathcal{V}$ to $v \in \mathcal{V}$ is a sequence $(u=k_0, k_1, \ldots, k_m=v)$ such that
$(k_{\ell-1},k_\ell)\in\mathcal{E}$ for all $\ell$; the nodes $k_1,\ldots,k_{m-1}$ are the internal nodes. Before we provide the general definition of the $\operatorname{Par}(\cdot)$, let us start with an example illustrating the reduction of the conditioning set using Markov property.

\paragraph{Illustrative example.}

\begin{figure}
\centering
\begin{tikzpicture}[
  every node/.style={circle, draw, minimum size=7mm, inner sep=0pt, font=\small},
  every edge/.style={draw}
]
% Node positions (a square for 1–4; node 5 attached to 4)
\node (v1) at (0,1) {1};
\node (v2) at (1,1) {2};
\node (v3) at (0,0) {3};
\node (v4) at (1,0) {4};
\node (v5) at (2,0) {5};

% Edges: (1,2),(1,3),(2,3),(3,4),(4,5)
\draw (v1) -- (v2);
\draw (v1) -- (v3);
\draw (v2) -- (v4);
\draw (v3) -- (v4);
\draw (v4) -- (v5);
\end{tikzpicture}
\caption{An example graph for illustrating the Markov property and the reduction of conditioning sets.}
\label{fig:example}
\vspace{-0.45cm}
\end{figure}
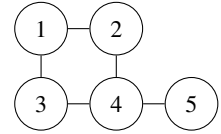

The construction of the parent nodes set is best exemplified through an example. Consider the undirected graph topology in \Cref{fig:example}, and consider the following permutation: $\sigma = (5,1,4,2,3).$
We construct each parent set $\operatorname{Par}(\sigma(i))$ incrementally by parsing previously ordered nodes $\sigma(i-1),\sigma(i-2),\ldots,\sigma(1)$ and adding a node only if it is reachable via a path whose internal nodes avoid the parents selected so far. We have $\operatorname{Par}(5)=\emptyset.$ For the second node $\sigma(2)=1$, since there is a path, for instance $1\!-3\!-\!4\!-\!5$ whose internal nodes $\{3,4\}$ are not in its parent set (initialized to the empty set), we add $5$ and obtain $\operatorname{Par}(1)=\{5\}.$
For node $\sigma(3)=4$, we have $\operatorname{Par}(4)=\{1,5\}$  since there are paths $4\!-\!3\!-\!1$ and $4\!-\!5$ where the internal node, $\{3\}$, is not in the parent set. Similarly, for nodes $\sigma(4)=2$ and $\sigma(5)=3$, we have $\operatorname{Par}(2)=\{1,4\}, \quad \operatorname{Par}(3)=\{1,4\}$.
Indeed, for both nodes $2$ and $3$, the set $\{5\}$ is screened by $\{4\}$, while $\{2\}$ and $\{3\}$ are screened by $\{1,4\}$, respectively. As a result, due to Markov property, the general decomposition in this example is reduced to the following simpler decomposition:
\begin{align*}
  p(x) = p(x_5) \, p(x_1 \mid x_5) \, p(x_4 \mid x_1,x_5) \\ p(x_2 \mid x_1,x_4) \, p(x_3 \mid x_1,x_4).
\end{align*}
Generalizing the example above, we use the following formal definition to reduce the number of nodes being conditioned on to a reduced $\operatorname{Par}(\cdot)$ set.
\begin{definition}
Fix a permutation $\sigma=(\sigma(1),\ldots,\sigma(N))$ and an undirected strongly connected graph $\mathcal{G}=(\mathcal{V},\mathcal{E})$.
\[
\operatorname{Par}(\sigma(1)) := \emptyset
\]
For each index $i\ge 2$, define the set of previously visited nodes
\[
V_{<i} := \{\sigma(1),\ldots,\sigma(i-1)\}.
\]
We define the parent set of $\sigma(i)$, $i\ge 2$ as
\begin{align*}
\operatorname{Par}(\sigma(i)) :=
\left\{
\begin{aligned}
    &\sigma(j)\in V_{<i} : \exists\ \text{a path } \pi \text{ from }\sigma(i)\text{ to} \\
    &\sigma(j) \text{ such that }\pi \cap \bigl(V_{<i}\setminus\{\sigma(j)\}\bigr)=\emptyset
\end{aligned}
\right\}.
\end{align*}

Equivalently, $\sigma(j)\in\operatorname{Par}(\sigma(i))$ if and only if
$\sigma(j)$ is reachable from $\sigma(i)$ in the induced subgraph obtained by
removing all previously visited nodes except $\sigma(j)$, i.e.,
in the graph $\mathcal{G}\bigl[\mathcal{V}\setminus (V_{<i}\setminus\{\sigma(j)\})\bigr]$.
\end{definition}
%%%%%%

\subsection{Reduced conditional distributions in the autoregressive decomposition}

For a node $k>1$, the number of terms in the parameterized conditional distribution \eqref{eq:CondProb} depends on the cardinality of its parent set $\operatorname{Par}(k)$, and the order of interactions. We reserve the symbol $|\cdot|$ to denote cardinality of a set. Suppose node $k$ has $d_k := |\operatorname{Par}(k)|$ parents, and we parameterize the conditional distribution using polynomial interactions of degree at most $O$. Each interaction term in the exponent \eqref{eq:CondProb} is of the form $x_k \prod_{j \in S} x_j$, where $S \subseteq \operatorname{Par}(k)$.
Restricting the total degree to be at most $O$ implies that $|S| \le O-1$.
Consequently, the number of interaction terms involving $x_k$
\[
T := 1+ \sum_{r=1}^{O-1} \binom{d_k}{r},
\]
where the $+1$ corresponds to the singleton interaction term corresponding to $S=\emptyset$. If however, $O-1 > d_k$, then the sum saturates to all subsets and then the number of terms equals $2^{d_k}$.
We define 
\[ S_k^{(O)} = \{S \subseteq \operatorname{Par}(k): 0 \le |S| \le O-1\}. \] The cardinality of this set is $T$.

For a $\sigma(i)$, the conditional distribution of order at most $O$, can therefore be expressed as
\begin{align}
\label{eq:CondProb2}
\begin{aligned}
& p\!\left( x_{\sigma(i)} \mid x_{\operatorname{Par}(\sigma(i))} \right) \\
& \quad= \frac{1}{Z_{\sigma(i)}} 
\exp \left( x_{\sigma(i)} \sum_{S \in \mathcal{S}_{\sigma(i)}^{(O)}} \tilde{\theta}_{\sigma(i),S} 
\prod_{j \in S} x_j \right),
\end{aligned}
\end{align}
where $\tilde{\theta}_{\sigma(i),S}$ denotes the effective interaction coefficient associated with the monomial
$x_{\sigma(i)} \prod_{j \in S} x_j$.

Due to the reduction to the parent sets, the general decomposition \eqref{eq:general_decomposition} is simplified to:
\begin{align}
\label{eq:sampling}
& p(x) = p(x_{\sigma(1)}) \; p\left(x_{\sigma(2)} | \operatorname{Par}(x_{\sigma(2)}\right) \nonumber \\
& \hspace{2mm} p\left(x_{\sigma(3)} | \operatorname{Par}(x_{\sigma(3)} \right) \ldots p\left(x_{\sigma(N)} | \operatorname{Par}(x_{\sigma(N)}\right).
\end{align}
\emph{using Markov property and without any approximation.}

Now that we have defined conditionals entering in the autoregressive decompositions, we will discuss methods for learning these conditional probabilities from data next.

% \textcolor{blue}{[Need to talk about RISE here.]}
% Interaction Screening Estimator (ISE) is a method for estimating the parameters $\theta_i, \theta_{i,j}$ of an Ising model given a sufficient number of samples. In this section, we introduce ISE. Suppose that there are $m$ total samples, denote the $l$\textsuperscript{th} sample by $ x^{(l)} = (x^{(l)}_1, \ldots, x^{(l)}_N)$. For each $i \in \mathcal{V}$, the ISE estimate for node $i$ is obtained by solving the following minimization problem. 
% % (\theta_i, \{\theta_{i,j}\}_{j \not=i} )
% \begin{align}
% \label{eq:ISE}
% \begin{aligned}
%         &\mathsf{S}_i = \frac{1}{m} \sum_{l=1}^{m} \exp \! \left( -\, \theta_i \, x^{(l)}_i \,-\, x^{(l)}_i \sum_{j \not=i} \theta_{i,j} 
%      x^{(l)}_j \right) \\
%         &\text{ISE}_i: \argmin_{(\theta_i,\theta_{i,j})} \Big[ \mathsf{S}_i + \lambda \Big( |\theta_i| + \sum_{j\not=i} |\theta_{i,j}| \Big) \Big]
%     \end{aligned}
% \end{align}
% where $\lambda>0$ is a regularization parameter.

\paragraph{Learning conditionals in the autoregressive decomposition.}

Given the choice of the permutation $\sigma$ and the construction of the parent sets resulting in the final simplified autoregressive decomposition \eqref{eq:sampling}, we discuss the method for learning of these conditionals from data. We use the unconstrained version of the method known as GRISE \cite{vuffray2020efficient} that learns conditional probabilities of discrete undirected graphical models with higher-order interactions \eqref{eq:CondProb2}. Suppose that we are given $m$ independent and identically distributed (i.i.d.) samples, and denote the $l$\textsuperscript{th} sample by $ x^{(l)} = (x^{(l)}_1, \ldots, x^{(l)}_N)$. The GRISE estimate of parameters of conditional distributions in \eqref{eq:CondProb2} for node $i$ is obtained by solving the following minimization problem:
\begin{align}
\label{eq:ISE2}
&\tilde{\mathsf{S}}_i = \frac{1}{m} \sum_{l=1}^{m} \exp\! \Big( -\tilde{\theta}_i \, x_i^{(l)} - x_i^{(l)} \sum_{S \in \mathcal{S}_i^{(O)}} \tilde{\theta}_{i,S} \prod_{j \in S} x_j^{(l)} \Big) ,\nonumber \\
&\text{GRISE}_i: \argmin_{ \tilde{\theta}_i, \{ \tilde{\theta}_{i,S}\}_{S \in \mathcal{S}_i^{(O)}} } \tilde{\mathsf{S}}_i .
\end{align}
We do not discuss the details of this method further, but we notice that parameters of conditionals of order $O$ can be consistently estimated in time $\mathcal{O}(N^{O})$.

\subsection{Selecting optimized ordering}
\label{sec:order_choice}

\paragraph{Learning of the Markov random field structure.} In the case where only data is initially available, and the graph $\mathcal{G}=(\mathcal{V},\mathcal{E})$ is unknown, its structure can be learned using existing methods. For this purpose, we use an exact and consistent method known as Regularized Interaction Screening Estimator (RISE) introduced in \cite{vuffray2016interaction} and with the hyper-parameters values prescribed in \cite{lokhov2018} without any change. Given that this Ising model graph learning step is standard, we do not discuss the details of this method here. In all numerical examples below, we first learn the undirected graphical model structure associated with the Markov random field, and then study the effect of different variable orderings.

\paragraph{Criterion for choosing optimized variable ordering.}
Once the MRF graph $\mathcal{G}=(\mathcal{V},\mathcal{E})$ is known or has been learned, we propose the following strategy for selecting the optimized variable ordering for autoregressive decompositions. For a given sequence $\sigma$, we use the procedure above to define the parent sets based on the MRF edge set $\mathcal{E}$. Denote $d := \max_k d_k = \max_k |\operatorname{Par}(k)|$ the maximum cardinality of the parent set in the autoregressive decomposition \eqref{eq:sampling}. Let's also denote $K$ the maximum number of conditionals with cardinality $d$ appearing in \eqref{eq:sampling}. According to the Theorem 2 in \cite{vuffray2020efficient}, these $K$ conditionals are the hardest to learn: the number of samples required to guarantee a fixed error on the conditionals of the type \eqref{eq:CondProb2} with degree $d$ scale exponentially in $d$. Conversely, given a fixed number of samples $m$, we expect the largest error to impact conditionals with the maximum interaction order $d$. Therefore, we put forward a natural hypothesis: given a fixed number of training samples $m$ and comparing two permutations, \emph{ (i) a permutation resulting in a decomposition \eqref{eq:sampling} with smaller $d$, and (ii) if $d$ are the same for both permutations, a decomposition with smaller $K$ } should result in more accurately learned conditionals and hence in a higher-quality autoregressive model. As a minor additional enhancement, choice of ordering can be further slightly optimized by selecting strongly correlated nodes and their parents in the lower-order conditionals, exploiting spatial correlation decay in maximum-order conditionals. We rigorously test this hypothesis under control settings of i.i.d. samples in small models in the next section.

%%%%%%%%%%%%%%%%%%%%%%%%%%%%%%%%%% TIKZ-specific %%%%%%%%%%%%%%%%%%%%%%%%%%%%%%%%%%
\definecolor{myred}{rgb}{0.792, 0.0, 0.125}
\definecolor{mypink}{rgb}{0.957,0.647,0.51}
\definecolor{myblue}{rgb}{0.02,0.443, 0.69}
%%%%%%%%%%%%%%%%%%%%%%%%%%%%%%%%%% SEQUENTIAL %%%%%%%%%%%%%%%%%%%%%%%%%%%%%%%%%%
\newcommand{\latticeSeq}{%
\begin{tikzpicture}[scale=0.66, transform shape, x=1cm, y=1cm,
  every node/.style={circle, draw, minimum size=7mm, inner sep=0pt, font=\scriptsize}
]

% --- Background layer setup (so the band goes behind nodes/edges) ---
\pgfdeclarelayer{background}
\pgfsetlayers{background,main}

\begin{pgfonlayer}{background}
  % parameters
  \def\opstart{0.70}    % opacity for first circle (k=0)
  \def\opend{0.10}      % opacity for last circle (k=24)
  \def\radius{0.45}     % circle radius (in grid units)

  \foreach \r in {0,...,4} {%
    \foreach \c in {0,...,4} {%
      % sequential index k in row-major order (0..24)
      \pgfmathtruncatemacro{\k}{\r*5+\c}%
      % linear interpolation of opacity: op = opstart - (opstart-opend)*(k/24)
      \pgfmathsetmacro{\op}{\opstart - (\opstart-\opend)*(\k/24)}%
      % draw circle at grid coordinate (\c,-\r)
      \fill[draw=none, fill=myred, opacity=\op] (\c,-\r) circle [radius=\radius];
    }%
  }%
\end{pgfonlayer}

% Define labels as a 5x5 table (edit these as you like)
\def\nodelist{1,2,3,4,5,6,7,8,9,10,11,12,13,14,15,16,17,18,19,20,21,22,23,24,25}

% Iterate labels in row-wise order and compute (r,c) from index
\foreach \lab [count=\k from 0] in \nodelist {%
  \pgfmathtruncatemacro{\r}{int(\k/5)}%
  \pgfmathtruncatemacro{\c}{mod(\k,5)}%
  \pgfmathtruncatemacro{\id}{\r*5+\c+1}%
  \node (v\id) at (\c,-\r) {\lab};
}
  % Horizontal edges
  \foreach \r in {0,...,4} {
    \foreach \c in {0,...,3} {
      \pgfmathtruncatemacro{\a}{\r*5+\c+1}
      \pgfmathtruncatemacro{\b}{\r*5+\c+2}
      \draw (v\a) -- (v\b);
    }
  }
  % Vertical edges
  \foreach \r in {0,...,3} {
    \foreach \c in {0,...,4} {
      \pgfmathtruncatemacro{\a}{\r*5+\c+1}
      \pgfmathtruncatemacro{\b}{(\r+1)*5+\c+1}
      \draw (v\a) -- (v\b);
    }
  }
\end{tikzpicture}%
}

%%%%%%%%%%%%%%%%%%%%%%%%%%%%%%%%%% SKIP %%%%%%%%%%%%%%%%%%%%%%%%%%%%%%%%%%
\newcommand{\latticeSkip}{%
\begin{tikzpicture}[scale=0.66, transform shape, x=1cm, y=1cm,
  every node/.style={circle, draw, minimum size=7mm, inner sep=0pt, font=\scriptsize}
]

% --- Background layer setup (so the band goes behind nodes/edges) ---
\pgfdeclarelayer{background}
\pgfsetlayers{background,main}

% --- Diagonal shaded transparent band (behind everything else) ---
\begin{pgfonlayer}{background}
    \fill[draw=none, fill=mypink, opacity=0.70] (0,0) circle [radius=0.45];
    \fill[draw=none, fill=mypink, opacity=0.65] (2,0) circle [radius=0.45];
    \fill[draw=none, fill=mypink, opacity=0.60] (4,0) circle [radius=0.45];
    \fill[draw=none, fill=mypink, opacity=0.55] (1,-1) circle [radius=0.45];
    \fill[draw=none, fill=mypink, opacity=0.50] (3,-1) circle [radius=0.45];
    \fill[draw=none, fill=mypink, opacity=0.45] (0,-2) circle [radius=0.45];
    \fill[draw=none, fill=mypink, opacity=0.40] (2,-2) circle [radius=0.45];
    \fill[draw=none, fill=mypink, opacity=0.35] (4,-2) circle [radius=0.45];
    \fill[draw=none, fill=mypink, opacity=0.30] (1,-3) circle [radius=0.45];
    \fill[draw=none, fill=mypink, opacity=0.25] (3,-3) circle [radius=0.45];
    \fill[draw=none, fill=mypink, opacity=0.20] (0,-4) circle [radius=0.45];
    \fill[draw=none, fill=mypink, opacity=0.15] (2,-4) circle [radius=0.45];
    \fill[draw=none, fill=mypink, opacity=0.10] (4,-4) circle [radius=0.45];
\end{pgfonlayer}

% Define labels as a 5x5 table (edit these as you like)
\def\nodelist{1,14,2,15,3,16,4,17,5,18,6,19,7,20,8,21,9,22,10,23,11,24,12,25,13}

% Iterate labels in row-wise order and compute (r,c) from index
\foreach \lab [count=\k from 0] in \nodelist {%
  \pgfmathtruncatemacro{\r}{int(\k/5)}%
  \pgfmathtruncatemacro{\c}{mod(\k,5)}%
  \pgfmathtruncatemacro{\id}{\r*5+\c+1}%
  \node (v\id) at (\c,-\r) {\lab};
}
  % Horizontal edges
  \foreach \r in {0,...,4} {
    \foreach \c in {0,...,3} {
      \pgfmathtruncatemacro{\a}{\r*5+\c+1}
      \pgfmathtruncatemacro{\b}{\r*5+\c+2}
      \draw (v\a) -- (v\b);
    }
  }
  % Vertical edges
  \foreach \r in {0,...,3} {
    \foreach \c in {0,...,4} {
      \pgfmathtruncatemacro{\a}{\r*5+\c+1}
      \pgfmathtruncatemacro{\b}{(\r+1)*5+\c+1}
      \draw (v\a) -- (v\b);
    }
  }
\end{tikzpicture}%
}

%%%%%%%%%%%%%%%%%%%%%%%%%%%%%%%%%% DIAG %%%%%%%%%%%%%%%%%%%%%%%%%%%%%%%%%%
\newcommand{\latticeDiag}{%
\begin{tikzpicture}[scale=0.66, transform shape, x=1cm, y=1cm,
  every node/.style={circle, draw, minimum size=7mm, inner sep=0pt, font=\scriptsize}
]

% --- Background layer setup (so the band goes behind nodes/edges) ---
\pgfdeclarelayer{background}
\pgfsetlayers{background,main}

% --- Shaded transparent band (behind everything else) ---
\begin{pgfonlayer}{background}
    \fill[draw=none, fill=myblue, opacity=0.60] (2,-2) circle [radius=0.45];
    
    \fill[draw=none, fill=myblue, opacity=0.45] (0,0) circle [radius=0.45];
    \fill[draw=none, fill=myblue, opacity=0.45] (4,-4) circle [radius=0.45];
    
    \fill[draw=none, fill=myblue, opacity=0.30] (1,-1) circle [radius=0.45];
    \fill[draw=none, fill=myblue, opacity=0.30] (3,-3) circle [radius=0.45];
    
    \fill[draw=none, fill=myblue, opacity=0.20] (2,0) circle [radius=0.45];
    \fill[draw=none, fill=myblue, opacity=0.20] (3,-1) circle [radius=0.45];
    \fill[draw=none, fill=myblue, opacity=0.20] (4,-2) circle [radius=0.45];
    \fill[draw=none, fill=myblue, opacity=0.20] (0,-2) circle [radius=0.45];
    \fill[draw=none, fill=myblue, opacity=0.20] (1,-3) circle [radius=0.45];
    \fill[draw=none, fill=myblue, opacity=0.20] (2,-4) circle [radius=0.45];
    
    \fill[draw=none, fill=myblue, opacity=0.10] (4,0) circle [radius=0.45];
    \fill[draw=none, fill=myblue, opacity=0.10] (0,-4) circle [radius=0.45];
\end{pgfonlayer}

% Define labels as a 5x5 table (edit these as you like)
\def\nodelist{2,14,6,22,12,18,4,15,7,23,9,19,1,16,8,24,10,20,5,17,13,25,11,21,3}

% Iterate labels in row-wise order and compute (r,c) from index
\foreach \lab [count=\k from 0] in \nodelist {%
  \pgfmathtruncatemacro{\r}{int(\k/5)}%
  \pgfmathtruncatemacro{\c}{mod(\k,5)}%
  \pgfmathtruncatemacro{\id}{\r*5+\c+1}%
  \node (v\id) at (\c,-\r) {\lab};
}

% Horizontal edges
\foreach \r in {0,...,4} {
\foreach \c in {0,...,3} {
  \pgfmathtruncatemacro{\a}{\r*5+\c+1}
  \pgfmathtruncatemacro{\b}{\r*5+\c+2}
  \draw (v\a) -- (v\b);
}
}
% Vertical edges
\foreach \r in {0,...,3} {
\foreach \c in {0,...,4} {
  \pgfmathtruncatemacro{\a}{\r*5+\c+1}
  \pgfmathtruncatemacro{\b}{(\r+1)*5+\c+1}
  \draw (v\a) -- (v\b);
}
}
\end{tikzpicture}%
}

%%%%%%%%%%%%%%%%%%%%%%%%%%%%%%%%%% Final %%%%%%%%%%%%%%%%%%%%%%%%%%%%%%%%%%
\begin{figure*}[!htb]
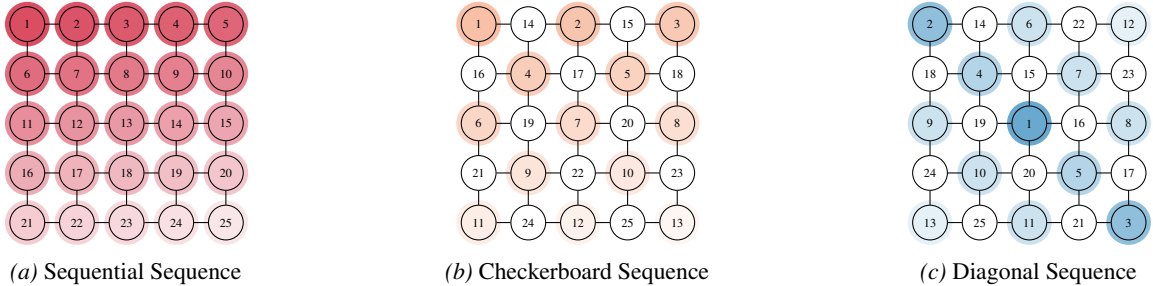

  \centering

  \begin{subfigure}[t]{0.3\textwidth}
    \centering
    \latticeSeq
    \caption{Sequential Sequence}
    \label{fig:lattice-seq}
  \end{subfigure}\hfill
  \begin{subfigure}[t]{0.3\textwidth}
    \centering
    \latticeSkip
    \caption{Checkerboard Sequence}
    \label{fig:lattice-skip}
      \end{subfigure}\hfill
  \begin{subfigure}[t]{0.3\textwidth}
    \centering
    \latticeDiag
    \caption{Diagonal Sequence}
    \label{fig:lattice-diag}
  \end{subfigure}
  \caption{Depiction of the three variable order traversals—sequential, checkerboard, and diagonal —on a $5\times 5$ lattice. Node shading indicates the relative selection order, from darker to lighter; identical shading denotes equal preference. Nodes without shading are selected after the shaded nodes and are not distinguished by the ordering. Node labels indicate one specific realization of the traversal.}
  \vspace{-0.3cm}
\end{figure*}
%%%%%%%%%%%%%%%%%%%%%%%%%%%%%%%%%%%%%%%%%%%%%%%%%%%%%%%%%%%%%%%%%%%%

\section{Results}
In this section, we validate our main hypothesis via numerical experiments. For the first two sets of experiments, we consider the two-spin Ising model on square lattices with lattice side length $L$, so that the total number of nodes is $N = L\times L$. We focus on smaller test models because we can guarantee that we produce high-quality i.i.d. training samples for these models. We then consider an application to real data produced by a quantum annealer. In all cases, we study two types of models.
\begin{enumerate}[nosep]
    \item \textbf{Ferromagnetic model:} describes a system of spins $x_i \in \{-1,+1\}$ with interactions favor alignment. We fix $\theta_i=0$ for all $i \in \mathcal{V}$ and $\theta_{i,j} = +1$, for $(i,j) \in \mathcal{E}$. Under this choice of parameters, the probability distribution~\eqref{eq:IsingMod} mostly concentrates on the two fully aligned configurations, namely $x_i = +1$ for all $i$ and $x_i = -1$ for all $i$.
    \item \textbf{Spin glass model}: describes a system of spins $x_i \in \{-1,+1\}$  with random couplings $\theta_{i,j} \in \{-1, 1\}$ or $\theta_{i,j} \in [-1,1]$, and with $\theta_i=0$. The randomness and competition between interactions lead to highly non-convex probability distribution \eqref{eq:IsingMod} with many probability peaks.
\end{enumerate}
% In the next few subsections, we present examples that will empirically show that some sequences have lower sampling error than others.
In all use cases (synthetic and real data), we first learn the MRF graph structure, and then use the learned structure to construct the structure-aware variable orderings. 

\subsection{Exact training samples: $5 \times 5$ square lattice}
In the first experiment, we consider a square lattice with $L=5$ ($N=25$). The reason for considering this lattice size is that i.i.d. samples can be produced by brute force for any Ising model (including spin glasses) and we don't have to worry about the quality of the training data in this setup. We fix the following three permutations $\sigma(1),\ldots,\sigma(N)$.

\begin{itemize}[nosep,leftmargin=*]
    \item \textbf{Sequence 1 (sequential order):} nodes are selected sequentially in row-wise order (see \Cref{fig:lattice-seq}). For a square lattice, this benchmark choice involves order $K=\mathcal{O}(L^2)$ of conditionals with the max cardinality $d=L$.
    \item \textbf{Sequence 2 (checkerboard order):} nodes are selected in the checkerboard pattern illustrated in \Cref{fig:lattice-skip}. For a square lattice, this benchmark choice still involves order $K=\mathcal{O}(L^2)$ of conditionals with the max cardinality $d=L$, but there are roughly twice less terms with the max cardinality (all remaining nodes in white have bounded cardinality 4 regardless of $L$), and the terms in each conditional should be less correlated with each other. We expect this choice to be better than the sequential benchmark. 
    % starting from node 1, this sequence is obtained by first selecting nodes such that, relative to each selected node, both the immediate neighbor to the right (in the same row) and the immediate neighbor below (in the next row) are skipped. We first select the nodes on one sublattice (e.g., those with $(r+c)$ even under row--column indexing), and then select the remaining nodes (see \Cref{fig:lattice-skip}).
    % \comr{Added: Choosing every other node decreases the correlation between nodes and their parent nodes, which ties with our hypothesis.... \textit{[we need to specify in the criterion about correlation.]}}
    \item \textbf{Sequence 3 (diagonal traversal):} this is the best optimized choice of variable ordering that we designed for a square lattice. The procedure is best understood by inspecting \Cref{fig:lattice-diag}. The main idea is that conditioning on of the main diagonals directly makes the two parts of the lattice conditionally independent (inside this diagonal, we also optimize the sequence to minimize correlations). Then, the graph can be partitioned further by choosing next nodes in the diagonals, skipping one diagonal at a time as these result in nodes with bounded cardinality 4 regardless of $L$. This strategy still unavoidably has terms with max cardinality $d=L$, but there are much fewer of them compared to the first two strategies.  
    % nodes are selected by traversing the lattice diagonally. We begin with the central node of the main diagonal (node 13 for \(L=5\)), and then visit nodes along the main diagonal in a zig-zag fashion, yielding the subsequence \(13,1,25,7,19\). We then move to the upper triangular region, skipping one diagonal at a time and selecting the next diagonal (e.g., \(3,9,15\)), before returning to the skipped diagonals (e.g., \(2,8,14,20\), followed by \(5\)). An analogous procedure is applied to the lower triangular region (see \Cref{fig:lattice-diag}). \comr{\textit{Added}: This sequence tries to satisfy both criteria of our hypothesis: decreasing $d_k$ and decreasing correlation between nodes and their parent set.}
\end{itemize}
The third permutation is constructed according to the hypothesis presented in \Cref{sec:order_choice} and serves as our test case; we compare its sampling error against the other two permutations that serve as baselines. We expect the test permutation to be optimized compared to the other two.

In realistic applications, learning of conditionals in the autoregressive decomposition \eqref{eq:sampling} will have two sources of error: (i) statistical error from finite number of training samples; (ii) systematic error from a reduced interaction order approximating the conditionals for large models. We use the small $N=25$ model to study the impact of each of these factors separately.
\begin{figure*}[!htb]
    \centering
    \subfloat[Ferromagnetic Model]{%
        \includegraphics[width=0.30\textwidth]
        {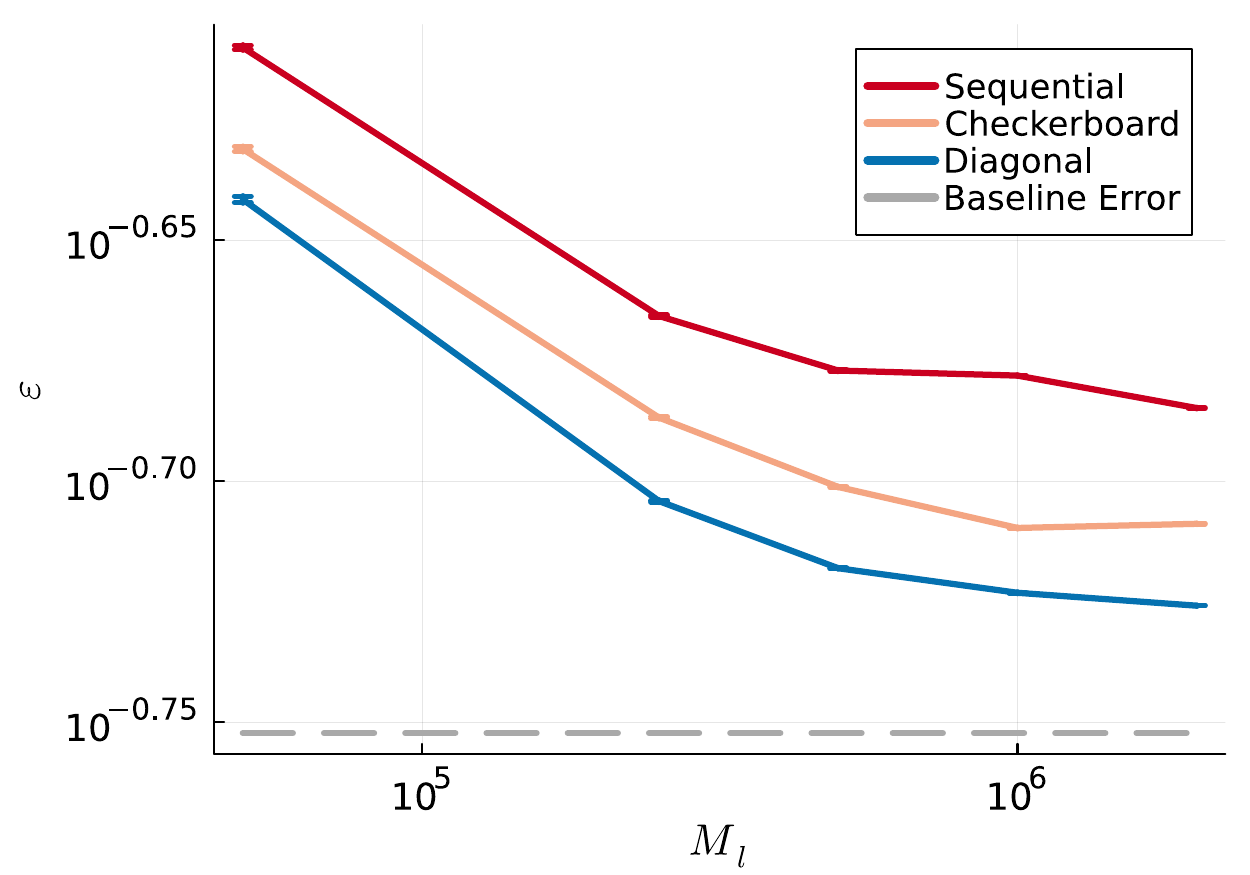}
        \label{subfig:ferro_5_hist}}%
    \hspace{1em}
    \subfloat[Spin glass Model]{%
        \includegraphics[width=0.30\textwidth]{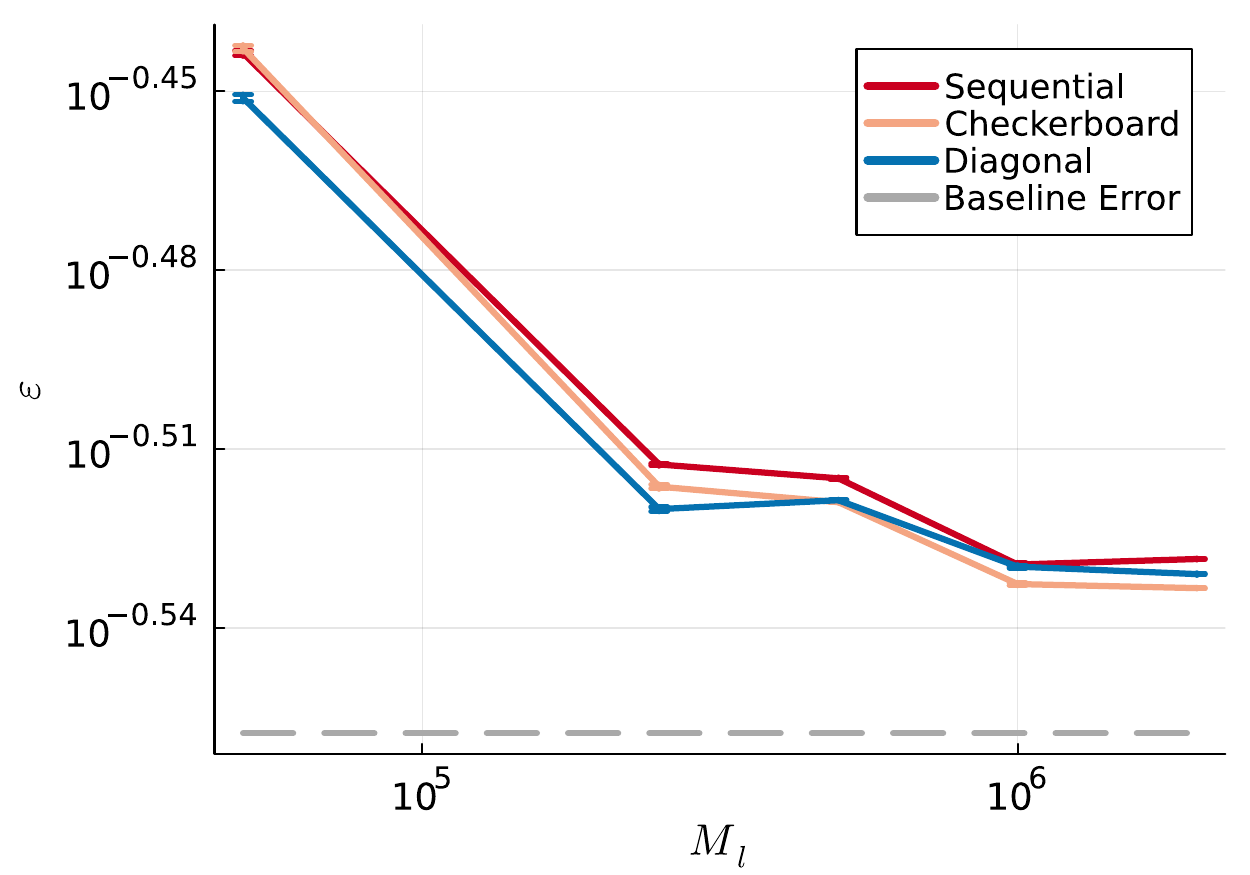}
        \label{subfig:spin_5_hist}}%
    \caption{Number of training data samples $M_l$ versus sampling error $\varepsilon$ \eqref{eq:error} for both ferro and spin glass Ising models with conditional order $O=6$. The number of generated samples $M_s = 10^5$. Error bars are one standard deviation over 20 independent training data sets.}
    \label{fig:5x5_hist}
    \vspace{-0.3cm}
\end{figure*}

\textbf{Learning models of fixed order using data samples.} In this set of experiments, we study how learning the conditional distributions in \eqref{eq:CondProb2} from a finite number of data samples affects the resulting sampling error.

\textit{Data Generation:} We generate 20 independently sampled datasets of size $M_l$ from a $5 \times 5$ square lattice for both the ferromagnetic and spin glass models. In the latter case, each dataset corresponds to a distinct instance of the model in which the edge couplings $\theta_{i,j} \in \{-1,1\}$ are drawn uniformly at random. We vary the number of training samples $M_l$ over the set $\{ 0.5 \times 10^5, ~ 2.5 \times 10^5, ~5 \times 10^5, ~10 \times 10^5, ~20 \times 10^5 \}$,

\textit{Learning Conditional Distributions:} The parameters in \eqref{eq:CondProb2} for each sequence are learned using the GRISE estimator defined in \eqref{eq:ISE2}. Since the underlying graph is a $5 \times 5$ square lattice, the maximum cardinality of $\operatorname{Par}(i)$ over all $i \in \mathcal{V}$ and for all three sequences considered is at most 5. Consequently, the order of the model, denoted by $O$ in \eqref{eq:CondProb2}, is fixed to $6$, accounting for a fully-expressible conditionals with interactions up to order six together with the bias term.

\textit{Sampling:} We fix the number of samples generated from the learned models to $M_s = 10^5$, using the sampling procedure described in \eqref{eq:sampling}. In \Cref{fig:5x5_hist}, we present $\log$-$\log$ plots of $M_l$ against the sampling error, defined to be the difference in the first two moments of the empirical distribution $\hat{p} := \frac{1}{M_s}\sum_{l=1}^{M_s} \delta_{x^{(l)}}$ and the true distribution $p$ \eqref{eq:IsingMod}:
\begin{align}
\label{eq:error}
\varepsilon =  (\left\| \mathbb{E}_{\hat{p}}[x] -\mathbb{E}_{p}[x] \right\|_2 + \left\| \operatorname{Cov}_{\hat{p}}(x) - \operatorname{Cov}_{p}(x) \right\|_F)^{\frac12},
\end{align}
where $\|\cdot\|_2, \|\cdot\|_F$ are the $2$-norm and the Frobenius norm, respectively. The vertical error bars represent the single error standard deviations over the 20 sampled datasets.

\textit{Discussion:} As shown in \Cref{fig:5x5_hist}, the sampling error decreases as $M_l$ increases, as expected, and saturates for larger values of $M_l$. To understand the saturation, we additionally plot the baseline sampling error arising solely from finite sampling, obtained by drawing $M_s$ samples directly from the ground-truth distribution \eqref{eq:IsingMod} (no model can produce better error than baseline due to finite samples). For the spin glass model, the baseline is averaged over the 20 random instances of the coupling realizations.

% The higher sampling error for the spin glass model is likely a consequence of the intrinsic disorder and frustration of the underlying distribution, which gives rise to a highly rugged energy landscape and complex, irregular correlations \cite{tahriri2025spin}. These features make the conditional distributions substantially harder to learn from finite data, leading to larger model error that dominates the sampling error.

In both model types, however, Sequence~3—the diagonal traversal—yields lower sampling error than the other two sequences; this improvement is particularly pronounced in the ferromagnetic case (\Cref{subfig:ferro_5_hist}). In contrast, the spin glass model (\Cref{subfig:spin_5_hist}) appears less sensitive to the choice of variable ordering, likely due to the hardness of sampling intrinsic to frustrated spin-glass models. Still, there is a clear separation (larger than the error bars) between the sampling error curves even in the spin-glass case.
% suggesting that unlike the ferromagnetic case, spin glass correlations do not exhibit a simple decay with lattice distance, reducing the effectiveness of structure-informed variable orderings.
% As a result, different traversal sequences induce conditionals of comparable complexity, and the sampling performance appears largely insensitive to the choice of ordering.

\textbf{Learning models of varying order using the true Distribution.} In this set of experiments, we study how order of the model, denoted by $O$ in \eqref{eq:CondProb2}, affects the sampling error in the ferromagnetic case. Moreover, the conditionals are learned from the true distribution, eliminating the error that stems from finite training samples.

\textit{Learning conditional distributions:} To isolate the effect of model order from errors arising due to finite training data, we learn the conditional distributions in \eqref{eq:CondProb2}, for each sequence, directly from the true probability distribution \eqref{eq:IsingMod}. Specifically, we enumerate all $2^{25}$ configurations $x$ and compute their exact probabilities $p(x)$, which are then used in place of empirical frequencies in the GRISE objective \eqref{eq:ISE2}. To learn the conditional distributions, we use model of orders that vary over $ O \in \{2, 4, 6\}$, and hence approximating the conditional probabilities in the decomposition \eqref{eq:sampling}. 

\begin{figure*}[!htb]
    \centering
    \subfloat[Order = 2]{%
        \includegraphics[width=0.30\textwidth]{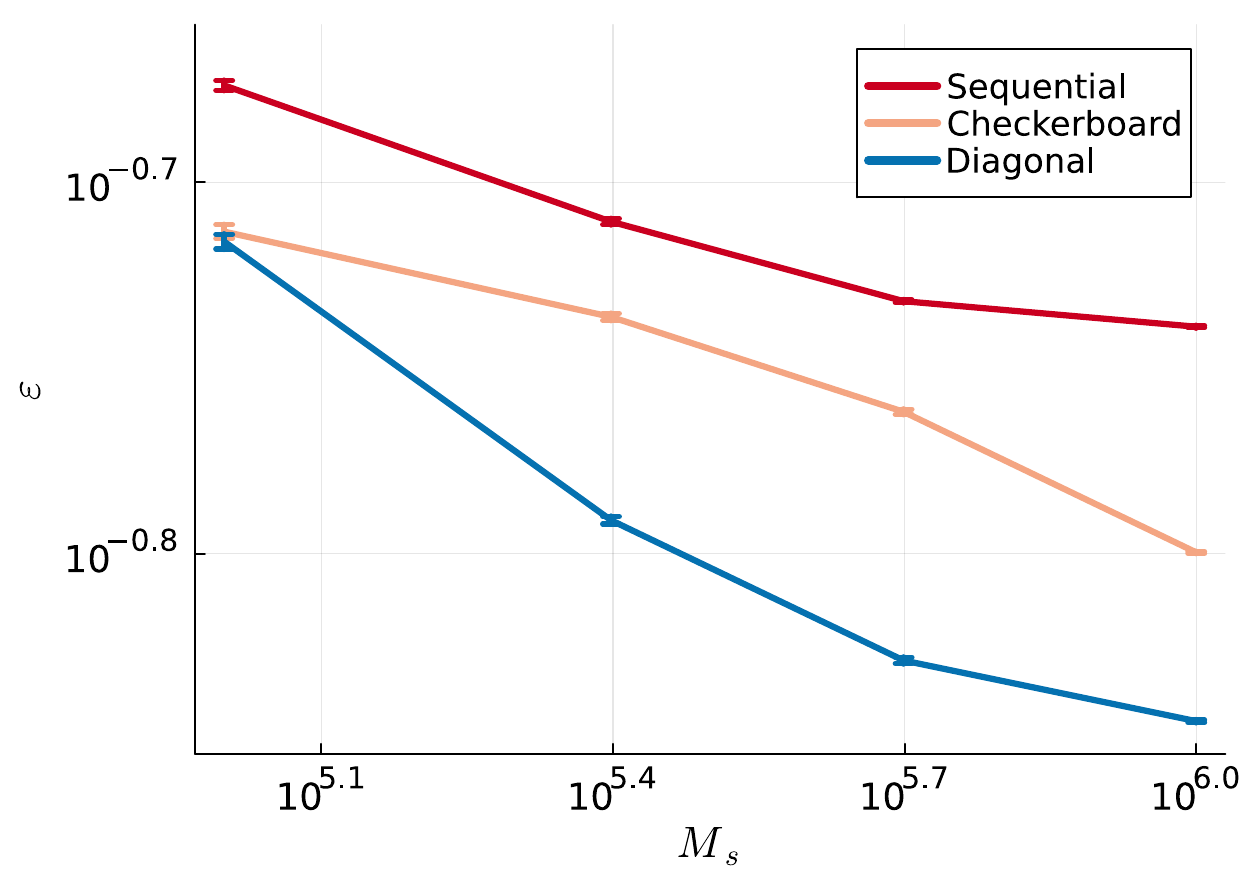}}%
    \hspace{1em}
    \subfloat[Order = 4]{%
        \includegraphics[width=0.30\textwidth]{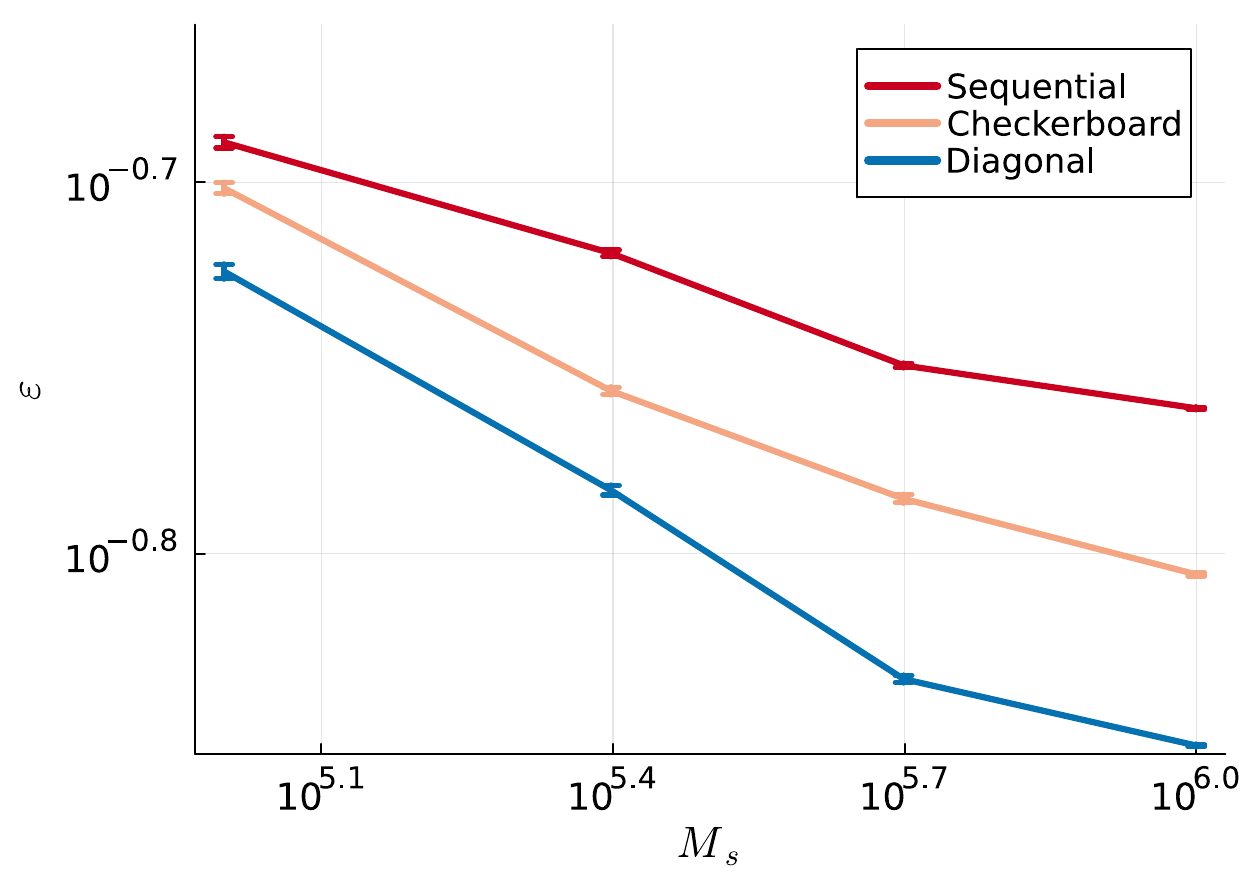}}%
    \hspace{1em}
    \subfloat[Order = 6]{%
        \includegraphics[width=0.30\textwidth]{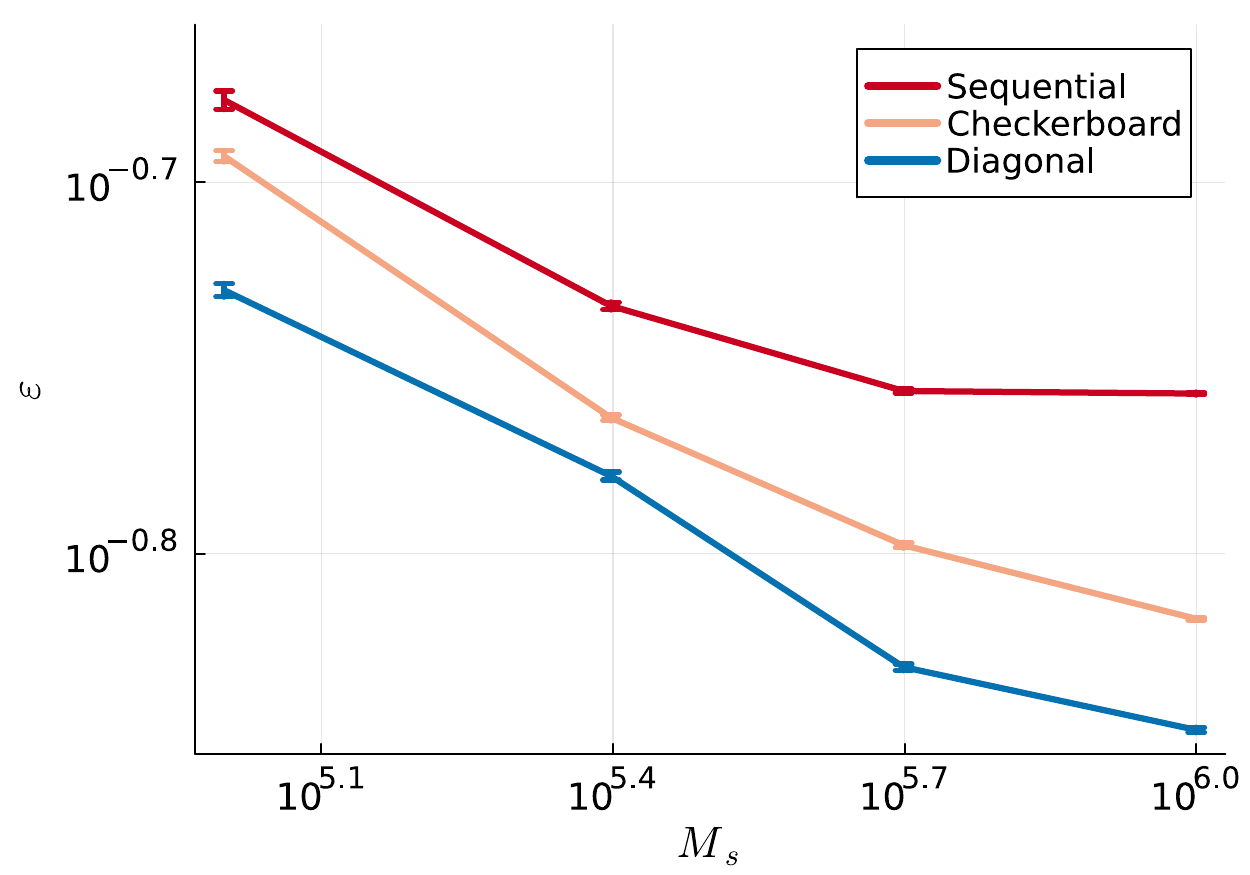}}%
    \caption{Number of generated samples $M_s$ versus sampling error $\varepsilon$ \eqref{eq:error} for a ferromagnetic Ising model with different conditional orders $O\in \{2,4,6\}$, where the conditionals are learned from the ground-truth distribution $p$. Error bars indicate one standard deviation over 50 independently generated sample sets.}
    \label{fig:5x5_true}
\end{figure*}

\textit{Sampling:} For each sequence and each model order, we generate $M_s$ samples from the learned autoregressive model, with $M_s \in \{ 1 \times 10^5, ~2.5\times10^5, ~5 \times 10^5, ~10 \times 10^5 \}$. In \Cref{fig:5x5_true}, we present the $\log$-$\log$ plots of $M_s$ verses the sampling error (the difference in the first two moments of the empirical and the true distribution). The reported results are averaged over 50 independent sampling runs, and the vertical error bars indicate the respective standard deviation. 

\textit{Discussion:} As shown in \Cref{fig:5x5_true}, the sampling error decreases with increasing  $M_s$ for all model orders. In all cases, Sequence~3 yields the lowest sampling error, validating our hypothesis. Interestingly, increase of the model order in learning of conditionals does not lead to a significant improvement in performance due to small value and impact of higher-order terms for the models of ferromagnetic type.
% \textcolor{blue}{[TODO: this is likely wrong, change]} \comr{\textit{Unsure but here is a reasoning}: The lack of a clear improvement with increasing model order indicates that low-order polynomial approximations are sufficient to represent the conditional distributions in this regime. This may be attributed to the relatively small system size, which limits the complexity of the effective conditional dependencies, reducing the impact of higher-order terms on sampling performance.}

\subsection{High-quality training samples: $10 \times 10$ ferro model}
We expect the effect already visible on models of size $N=25$ to be pronounced in larger models as well. It is however harder to perform numerical experiments under controlled setting as a reliable production of i.i.d. samples becomes an issue. Luckily, high-quality samples can still be obtained for specific models such as ferromagnetic models without local field. This is precisely the type of models that we study in this second experiment. We consider the same three permutations as in the previous section, and study the combined effect of the reduced model order in learning of conditionals (due to computational complexity) and the finite number of training samples on the sampling error.
% In the second experiment, we consider a larger square lattice. As in the previous case, we examine three permutations: Sequence~1, defined in a row-wise manner; Sequence~2, defined using a checkerboard ordering; and Sequence~3, defined via a diagonal traversal of the graph topology. In this setting, we study the combined effect of the model order and the finite number of training samples on the sampling error, focusing exclusively on the ferromagnetic model.

\textit{Data generation.}
% For smaller systems, data samples can be generated directly using Julia’s \texttt{StatsBase.sample}, which requires explicit enumeration of all configurations and their probabilities. For a $10 \times 10$ lattice, however, the configuration space grows to $2^{100}$, making this approach infeasible.
We generate data using a Markov chain Monte Carlo algorithm known as Gibbs sampling. For this specific model, we can use a special trick to avoid issues with convergence of Markov chains based on the fact that we know two configurations maximizing the probability density. We thus run two independent Markov chains initialized at $x = (1,1,\ldots,1)$ and $x = (-1,-1,\ldots,-1)$, discard an initial burn-in period, and combine
an equal number of samples from each chain to form the final dataset. Training datasets of size $M_l \in \{0.5 \times 10^5,\; 2.5 \times 10^5,\; 5 \times 10^5\}$ are generated. In the ferromagnetic case, where the distribution is dominated by two magnetized modes, this procedure ensures that both modes are adequately sampled and eliminates bias due to slow mixing between them.

\textit{Learning conditional distributions.}
For each sequence, we learn the parameters of the conditional model~\eqref{eq:CondProb2} using the Interaction Screening Estimator~\eqref{eq:ISE2}. In this experiment, we fix the model order to $O=2$ and $O=4$.

\begin{figure}[!htb]
    \centering
    \subfloat[Order = 2]{%
        \includegraphics[width=0.39\textwidth]{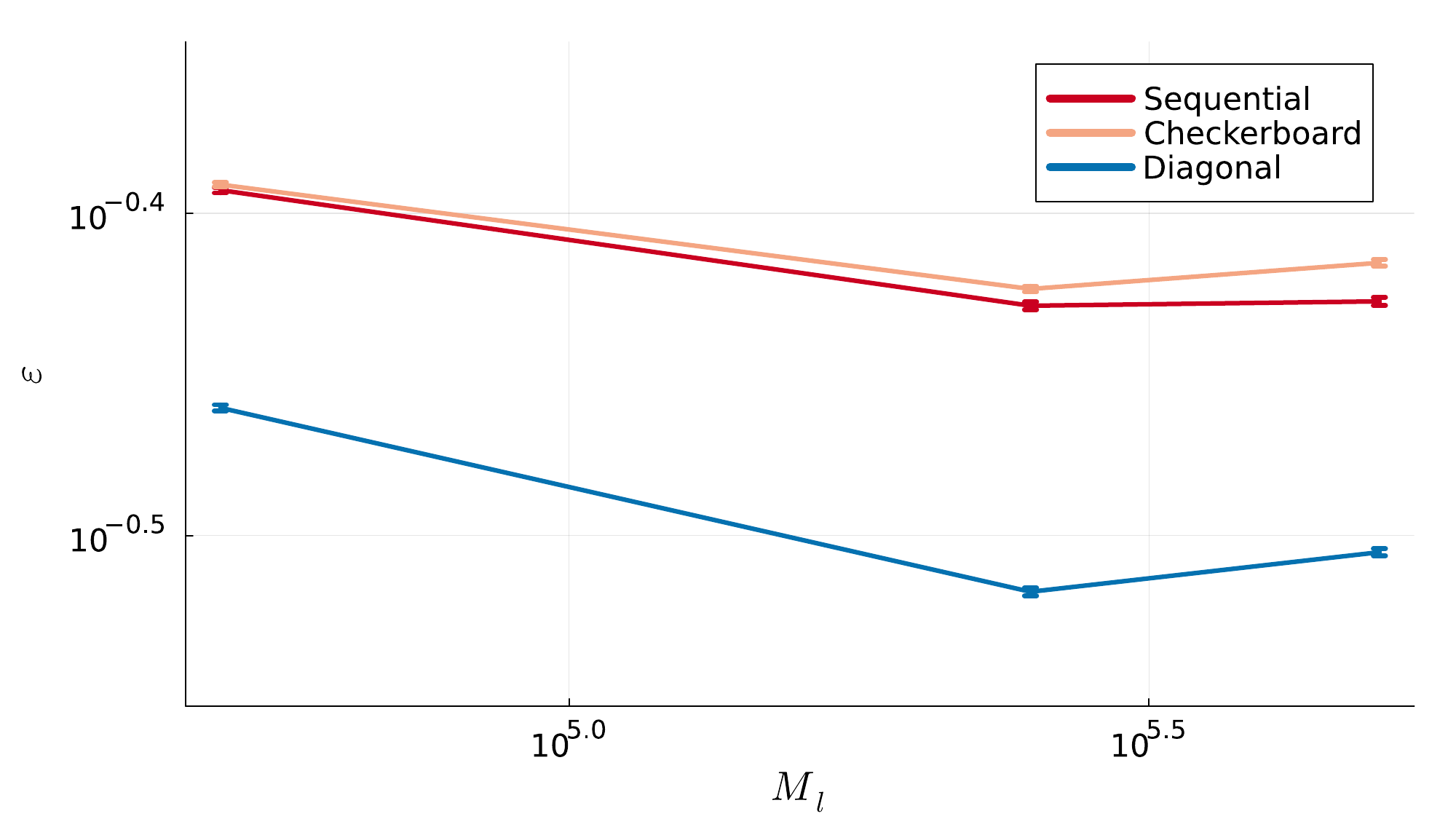}}%
    \hspace{1em}
    \subfloat[Order = 4]{%
        \includegraphics[width=0.39\textwidth]{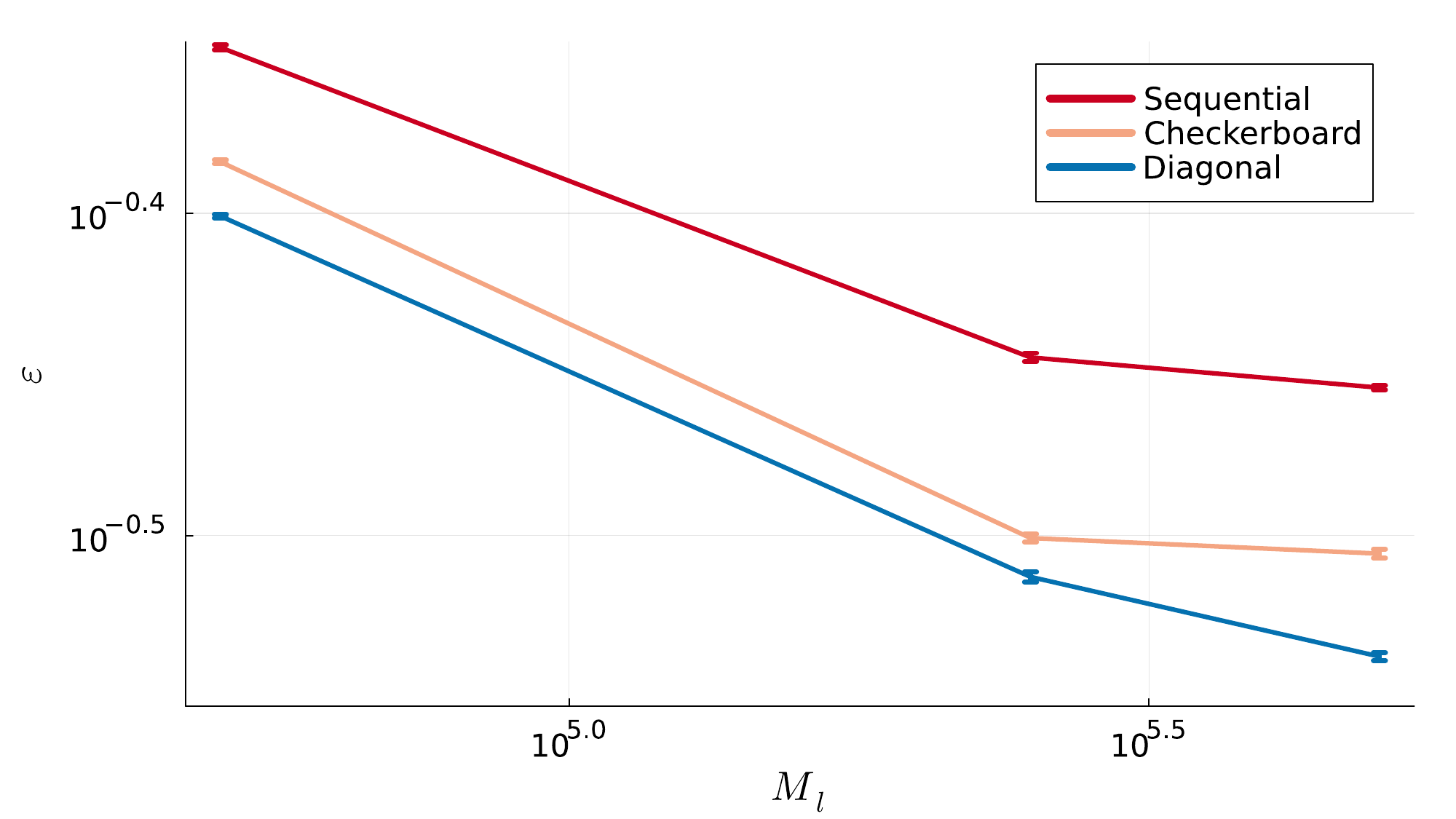}}%
    \caption{Number of training data samples $M_l$ versus sampling error $\varepsilon$ \eqref{eq:error} for ferro Ising model with model orders $O \in\{2,4\}$. The number of generated samples $M_s = 10^5$. Error bars indicate one standard deviation over 20 independent training data sets.}
    \label{fig:10x10_hist}
    \vspace{-0.39cm}
\end{figure}

\textit{Sampling.}
The number of generated samples is fixed to $M_s = 10^5$. In \Cref{fig:10x10_hist}, we report $\log$-$\log$ plots of the sampling error as a function of $M_l$. The sampling error is averaged over 50 independently generated sample sets, and error bars correspond to one standard deviation.

\textit{Discussion.}
As shown in \Cref{fig:10x10_hist}, the model order in~\eqref{eq:CondProb2} plays a more pronounced role in this larger system. Compared to the $5 \times 5$ lattice, the increased system size leads to more complex conditional dependencies, making the expressivity of the conditional model more consequential. In particular, the sampling error saturates more rapidly for the lower-order model ($O=2$) than for the higher-order model ($O=4$), suggesting that the lower-order model becomes capacity-limited at smaller training sizes. In contrast, the higher-order model continues to benefit from additional training data. Moreover, across both model orders, Sequence~3 consistently yields the lowest sampling error, highlighting the advantage of structure-informed diagonal traversals in larger, ordered lattices.

\subsection{Real data use case: D-Wave Dataset}

\begin{figure}[t]
    \centering
    \subfloat[Sequential Sequence]{%
        \includegraphics[width=0.32\columnwidth]{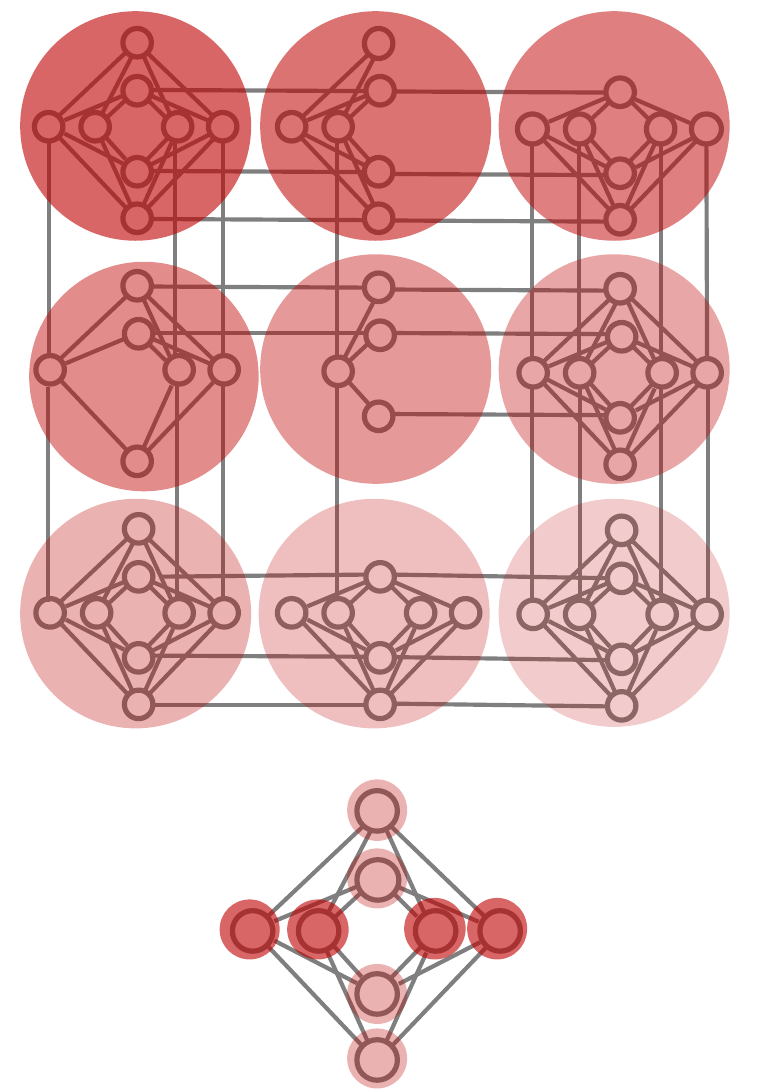}
        \label{fig:D-wave-seq}}%
    \hspace{1em}
    \subfloat[Cross Sequence]{%
        \includegraphics[width=0.32\columnwidth]{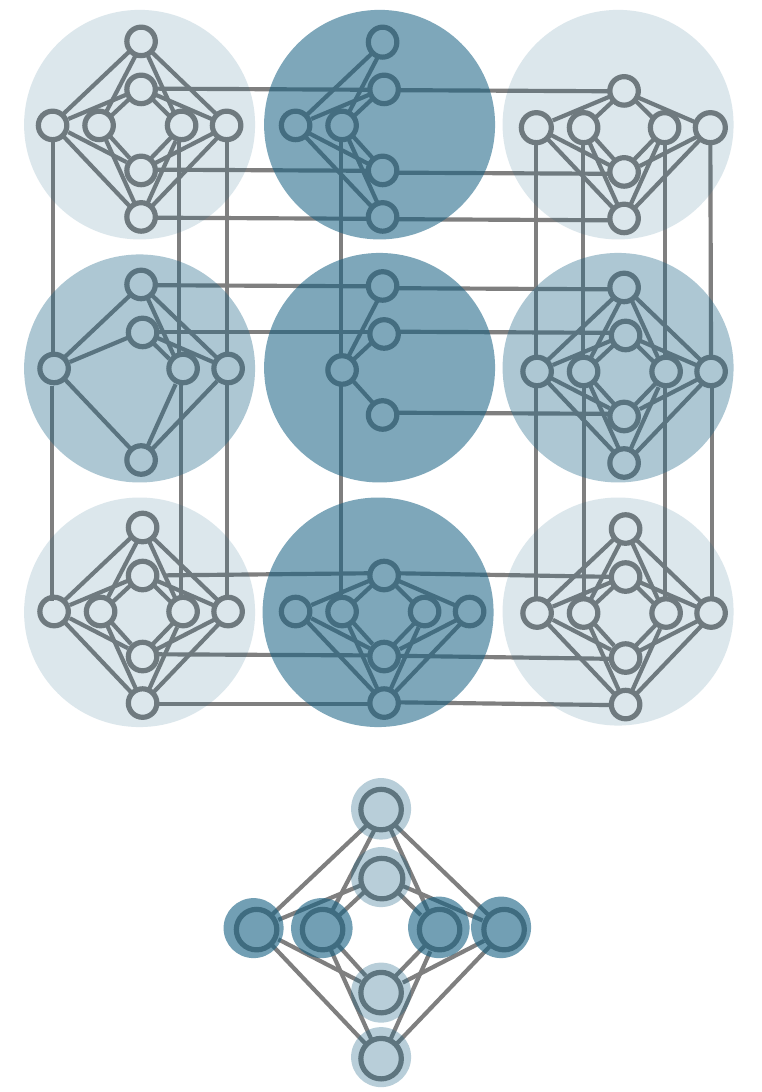}
        \label{fig:D-wave-cros}}%
    \caption{Depiction of two sampling sequences for the D-Wave dataset. The top panel illustrates the sublattice-level selection, while the bottom panel shows the selection order of nodes within each sublattice. Node shading indicates selection order.}
    \label{fig:D-wave-chip}
    \vspace{-0.2cm}
\end{figure}

\begin{figure}[t]
    \centering
    \includegraphics[width=0.39\textwidth]{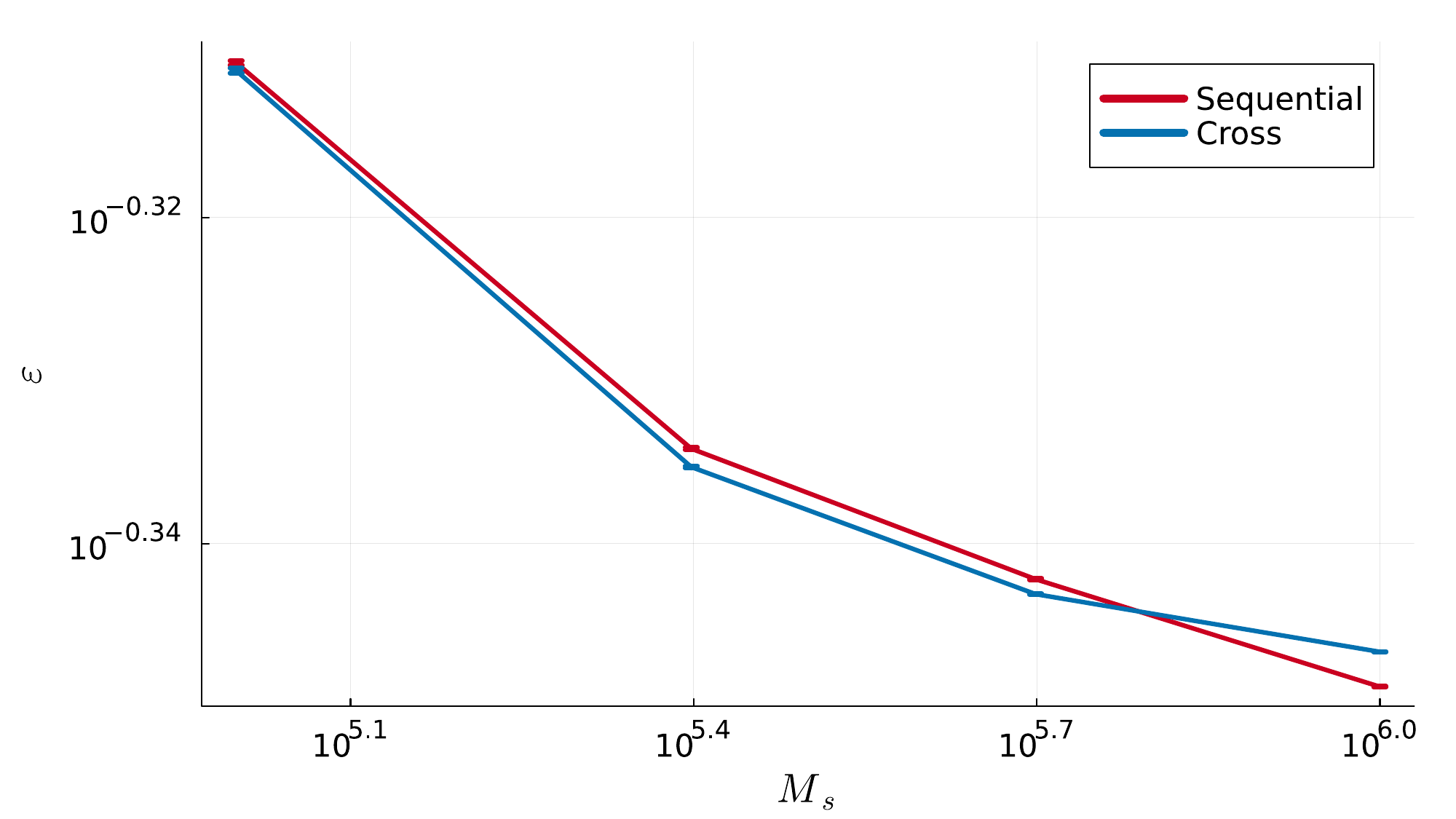}
    \caption{Number of generated samples $M_s$ versus sampling error $\varepsilon$ \eqref{eq:error} for the D-Wave spin glass Ising model with model order $O=3$, where the conditional distributions are learned from $M_l = 5 \times 10^5$. Error bars indicate one standard deviation over 20 independently generated sample sets.}
    \label{fig:D-wave}
    \vspace{-0.39cm}
\end{figure}
In this experiment, we evaluate the effect of variable permutations on the sampling algorithm in a non-synthetic setting, using data produced by a D-Wave 2X quantum annealer. The publicly available dataset \cite{lokhov2018} consists of recorded samples of $62$ qubits (i.e., $|\mathcal{V}|=62$) together with their empirical frequencies. The qubits are arranged on a non-square lattice, see \Cref{fig:D-wave-chip} the connectivity can be viewed approximately as a $9 \times 9$ lattice in which each node is itself replaced by a small, sparsely connected subgraph resembling a $4 \times 2$ lattice; several qubits are broken, represented by the missing edges in see Figure~\ref{fig:D-wave-chip}, which represents another interesting aspect of this dataset due to a connectivity irregularity.

The underlying probability distribution is unknown. However, the system is known to be well approximated by an Ising model with both pairwise couplings $\theta_{i,j}$ and local fields $\theta_i$, with parameters drawn uniformly from the range $[-2.0,-0.25] \cup [0.25,2.0]$. Consequently, this instance corresponds to a spin glass model.

\textit{Learning the graph topology.} \textit{Learning the graph topology.} We first reconstruct the underlying MRF graph topology as explained in the methods section. Based on the recovered edge weights, we then consider the following two sampling sequences: sequential depicted in \Cref{fig:D-wave-seq} (similar to the baseline used in square lattice experiments), and cross order depicted in \Cref{fig:D-wave-cros} (similar to the optimized variable order in the synthetic experiments).
% \begin{itemize}[nosep]
%     \item \textbf{Sequence 1 (sequential order):} nodes are selected sequentially, first the nodes in the row, followed by the nodes in the column within each $4 \times 2$ subgraph, and then traversing these subgraphs column-wise across the $9 \times 9$ lattice (see \Cref{fig:D-wave-seq}).
%     \item \textbf{Sequence 2 (cross order):} at the level of the $9 \times 9$ lattice, we first select the nodes in the central column, followed by those in the central row, and then the nodes located in the four corner regions. Within each $4 \times 2$ subgraph, similar to the previous case, nodes are selected sequentially, first along the row, followed by the nodes along the column (see \Cref{fig:D-wave-cros}). 
% \end{itemize}

\textit{Learning conditional distributions.}
We fix the model order to $O=3$, and learn the corresponding conditional distributions for both sequences using the Interaction Screening Estimator~\eqref{eq:ISE2} and the available data samples.

\textit{Sampling.} For each learned model, we generate $M_s$ samples with
\( M_s \in \{1 \times 10^5,\; 2.5 \times 10^5,\; 5 \times 10^5,\; 10 \times 10^5\}.\) We average results over 50 independent sampling runs, and use the same error metric corresponding to two moments of the distribution.
% To quantify the sampling error, we compute the first two moments of the empirical distribution induced by the generated samples and compare them against the corresponding moments estimated from the training dataset. In \Cref{fig:D-wave}, we present a $\log$-$\log$ plot of $M_s$ versus the sampling error. The reported results are averaged over 50 independent sampling runs, and the error bars regions indicate the corresponding variance.

\textit{Discussion.}
Results in \Cref{fig:D-wave} show a similar trend to spin-glass model in the synthetic experiments: the sampling error decreases with increasing $M_s$, as expected, and small sensitivity to the choice of variable ordering. This behavior is consistent with the strongly disordered nature of the D-Wave dataset. Still, it is interesting to see that the structure-aware cross order consistently shows an advantage over the naive ordering, supporting the general finding of this work. 

% As shown in \Cref{fig:D-wave}, the sampling error decreases with increasing $M_s$, as expected. However, no statistically significant difference is observed between the two traversal sequences. This behavior is consistent with the strongly disordered nature of the D-Wave dataset, which corresponds to a spin glass model with random couplings and local fields. In such systems, correlations are highly irregular and are not well aligned with geometric locality in the underlying lattice, limiting the extent to which structure-informed permutations can simplify the conditional distributions. Consequently, different variable orderings induce conditional models of comparable complexity, and the resulting sampling performance is largely insensitive to the choice of traversal sequence.

% \section{neuRISE}
% \[
% \frac1m \sum_{l=1}^m \exp( -\langle\Phi_l(x_i^{(l)},\operatorname{NN}(\operatorname{Par}(x_i^{(l)}),w) \rangle
% \]
% where the $\Phi_a$, for $a$ ranging over the spins $\{-1,1\}$, are the centered indicator basis defined as
% \[
% \Phi_a(x_i) = \begin{cases}
%     1-\frac1q, & \text{if } a = x_i \\
%     -\frac1q, & \text{otherwise}
% \end{cases}
% \]

\section{Conclusions and future work}
In this work, we deliberately focused on small-scale models where exact or well-controlled sampling procedures are available, allowing us to isolate and analyze the effect of variable ordering in autoregressive sampling. Even in these small models, we observe clear and systematic differences in sampling performance across orderings. These effects are expected to become more pronounced for larger systems, where conditioning complexity and error accumulation grow with system size. We also restricted our study to explicit parametric forms in order to quantify the impact of model expressivity on the performance of models with different variable orderings. Future work should benchmark the conclusions from this work on large models and datasets, using neural-net representations for the conditionals \cite{jayakumar2020learning} and exploring the extension to continuous variables. 

The code reproducing all results in this work is available at  \href{https://github.com/lanl-ansi/ARmodelVariableOrder}{Github}.

% \section*{Acknowledgements}

% \textbf{Do not} include acknowledgements in the initial version of the paper

% If a paper is accepted, the final camera-ready version can (and usually should)
% include acknowledgements.  Such acknowledgements should be placed at the end of
% the section, in an unnumbered section that does not count towards the paper
% page limit. Typically, this will include thanks to reviewers who gave useful
% comments, to colleagues who contributed to the ideas, and to funding agencies
% and corporate sponsors that provided financial support.

% \section*{Code Availability}
% The code reproducing all results in this work is available at  \href{https://anonymous.4open.science/r/Sampling-using-AR-models-A77A/README.md}{Anonymous Github}.

% \section*{Impact Statement}
% This paper presents work whose goal is to advance the field of Machine
% Learning. There are many potential societal consequences of our work, none
% which we feel must be specifically highlighted here.
\subsection*{Acknowledgment}
This work has been supported by the U.S. Department of Energy/Office of Science Advanced Scientific Computing Research Program.

\bibliography{literature}
\bibliographystyle{icml2026}

%%%%%%%%%%%%%%%%%%%%%%%%%%%%%%%%%%%%%%%%%%%%%%%%%%%%%%%%%%%%%%%%%%%%%%%%%%%%%%%
%%%%%%%%%%%%%%%%%%%%%%%%%%%%%%%%%%%%%%%%%%%%%%%%%%%%%%%%%%%%%%%%%%%%%%%%%%%%%%%
% APPENDIX
%%%%%%%%%%%%%%%%%%%%%%%%%%%%%%%%%%%%%%%%%%%%%%%%%%%%%%%%%%%%%%%%%%%%%%%%%%%%%%%
%%%%%%%%%%%%%%%%%%%%%%%%%%%%%%%%%%%%%%%%%%%%%%%%%%%%%%%%%%%%%%%%%%%%%%%%%%%%%%%
% \newpage
% \appendix
% \onecolumn

%%%%%%%%%%%%%%%%%%%%%%%%%%%%%%%%%%%%%%%%%%%%%%%%%%%%%%%%%%%%%%%%%%%%%%%%%%%%%%%
%%%%%%%%%%%%%%%%%%%%%%%%%%%%%%%%%%%%%%%%%%%%%%%%%%%%%%%%%%%%%%%%%%%%%%%%%%%%%%%

\end{document}